\documentclass[11pt]{article}

\usepackage{acl}
\usepackage{multirow}
\usepackage{times}
\usepackage{latexsym}
\usepackage{amsmath}
\usepackage{graphicx}

\usepackage[T1]{fontenc}

\usepackage[utf8]{inputenc}

\usepackage{microtype}

\usepackage{inconsolata}

\usepackage{graphicx}

\title{Gradient Masters at BLP-2025 Task 1: Advancing Low-Resource NLP for Bengali using Ensemble-Based Adversarial Training for Hate Speech Detection}

\author{
  Syed Mohaiminul Hoque$^*$ \and Naimur Rahman$^*$ \and Md Sakhawat Hossain$^*$ \\[2mm]
  \texttt{syedmuhaimintahsin@gmail.com} \\
  \texttt{naimur79@student.sust.edu} \\
  \texttt{sakhawatdhrubo@gmail.com}
}

\usepackage{amsmath}
\begin{document}
\maketitle
\begin{abstract}
This paper introduces the approach of "Gradient Masters" for BLP-2025 Task 1: "Bangla Multitask Hate Speech Identification Shared Task". We present an ensemble-based fine-tuning strategy for addressing subtasks 1A (hate-type classification) and 1B (target group classification) in YouTube comments. We propose a hybrid approach on a Bangla Language Model, which outperformed the baseline models and secured the 6th position in subtask 1A with a micro F1 score of 73.23\% and the third position in subtask 1B with 73.28\%. We conducted extensive experiments that evaluated the robustness of the model throughout the development and evaluation phases, including comparisons with other Language Model variants, to measure generalization in low-resource Bangla hate speech scenarios and data set coverage. In addition, we provide a detailed analysis of our findings, exploring misclassification patterns in the detection of hate speech.\footnote{Our code and models are available at \url{https://github.com/SyedT1/Shared_Task1_HateSpeech}}
\end{abstract}




\section{Introduction}
Both positive discourse and harmful language are amplified by social media where automatic detection of hate speech is crucial. For low-resource languages such as Bangla, practical moderation is made harder by limited labeled data, extreme class imbalance, and noisy user text which includes mixed scripts, transliteration, and frequent typos. The BLP-2025 Task 1 provides a challenging, fine-grained annotation scheme over YouTube comments comprising both (a) hate-type categories (e.g., Abusive, Political Hate, Sexism) and (b) target groups (e.g., Individual, Organization, Community).

In this work we try to present an ensemble-based fine-tuning pipeline designed to improve robustness and generalization on this task. Our design choices respond to two task-specific challenges:
\begin{itemize}
  \item \textbf{Noise \& Orthography:} Bangla social text shows spelling variation and mixed scripts. To tackle this we apply a lightweight rule-based normalization to reduce orthographic variance. before tokenization.
  \item \textbf{Adversarial/noisy input:} User comments often contain typos, slang, or misspellings. In order to improve model resilience, we apply adversarial fine-tuning using FGSM, perturbing token embeddings during training which simulates realistic input variations. 
\end{itemize}

 Our contributions are:
\begin{enumerate}
  \item A reproducible ensemble pipeline for robust Bangla hate speech detection.
  \item Empirical analysis of adversarial fine-tuning and normalization effects under severe class imbalance.
  \item Error analysis and per-class metrics that identify concrete failure modes for minority categories.
\end{enumerate}

\section{Related works}

Recent work on hate speech detection in low resource languages has evolved significantly. \cite{ghosh2022hate} analyze transformer-based monolingual and cross-lingual models (e.g., BERT, ALBERT, RoBERTa, DistilBERT, MuRIL) on datasets in Hindi, Marathi, Bangla, Assamese, and new Assamese and Bodo corpora, showing multilingual models generalize better to unseen low-resource languages, though performance varies by language family (Assamese–Bodo vs. Indo-Aryan). \cite{das2024survey} surveys hate speech detection in low-resource languages worldwide, cataloging datasets, features, and methods, highlighting dataset scarcity and linguistic variation as key barriers. In code-mixed settings, \cite{c-v-etal-2024-faux} introduce a multitask framework with XLM-RoBERTa for misinformation and hate speech in Hindi–English posts, demonstrating effectiveness of cross-lingual transformers with tweaks. For Bangla, \cite{AlMarufEtAl2024} survey linguistic nuances (e.g., dialect, script, informal speech) and challenges in hate speech detection. \cite{faria2024llms} compare LLMs (GPT-3.5 Turbo, Gemini 1.5 Pro) with traditional classifiers in Bangla via zero- and few-shot learning, achieving gains over SVM baselines. New datasets like BIDWESH \cite{fayaz2025bidwesh} and BanTH \cite{haider2025banth} provide multi-dialect or transliterated Bangla corpora with fine-grained labels for hate presence, type, and targets (e.g., gender, religion, origin). Earlier Bangla datasets by \cite{karim2020benchmarks, romim2021hate, romim2022bdshs} offer 30–50K comments for binary or categorical classification. BLP-2023 shared tasks \cite{raihan2023blp, tariquzzaman2023blp, fahim2023blp} advance robust systems: \cite{raihan2023blp} use a two-step pipeline for Non-Violence, Passive-Violence, and Direct-Violence with back-translation and multilinguality; \cite{tariquzzaman2023blp} combine Informal Bangla FastText embeddings with BiLSTM for cost-effective performance near transformers; \cite{fahim2023blp} enhance BanglaBERT via in-task pretraining, adversarial perturbation, and ensembles for sentiment, applicable to multi-task hate classification. Our work builds on these by integrating type, target, and severity in Bangla YouTube comments, emphasizing fine-grained labels and efficient models for low-resource settings.

\section{Dataset Description}

The dataset for this task \cite{hasan2025llm, blp2025-overview-task1}   exhibits significant class imbalance across both subtasks, with the "None" label constituting the majority in both training and development sets. Table \ref{tab:subtask1a-distribution} presents the distribution for Subtask 1A (Type of Hate) and Table \ref{tab:subtask1b-distribution} shows the distribution for Subtask 1B (Target Group).

\makeatletter
\newcommand{\commonsplitfootnote}{%
  \textbf{Dev} – public development set (hyper-parameter tuning); \\
  \textbf{Dev Test} – blind validation set (used for leaderboard); \\
  \textbf{Test} – official blind test set (final ranking). \\
  \textit{This footnote applies to both Table~\ref{tab:subtask1a-distribution} and Table~\ref{tab:subtask1b-distribution}.}}
\makeatother

\begin{table}[h]
\centering
\resizebox{\linewidth}{!}{%
\begin{tabular}{lrrrr}
\hline
\textbf{Label} & \textbf{Train} & \textbf{Dev} & \textbf{Dev Test} & \textbf{Test}\\
\hline
None & 19,954 & 1,451 & 1,447 & 5,751\\
Abusive & 8,212 & 564 & 549 & 2,312\\
Political Hate & 4,227 & 291 & 283 & 1,220\\
Profane & 2,331 & 157 & 185 & 709\\
Religious Hate & 676 & 38 & 40 & 179\\
Sexism & 122 & 11 &  8 & 29\\
\hline
\textbf{Total} & 35,522 & 2,512 & 2,512 & 10,200\\
\hline
\end{tabular}}
\caption{Subtask 1A (Type of Hate) Dataset Distribution}
\label{tab:subtask1a-distribution}
\end{table}

\begin{table}[h]
\centering
\resizebox{\linewidth}{!}{%
\begin{tabular}{lrrrr}
\hline
\textbf{Label} & \textbf{Train} & \textbf{Dev} & \textbf{Dev Test} & \textbf{Test}\\
\hline
None & 21,190 & 1,536 & 1,528 & 6,093\\
Individual & 5,646 & 364 & 391 & 1,571\\
Organization & 3,846 & 292 & 292 & 1,152\\
Community & 2,635 & 179 & 159 & 759\\
Society & 2,205 & 141 & 142 & 625\\
\hline
\textbf{Total} & 35,522 & 2,512 & 2,512 & 10,200\\
\hline
\end{tabular}}
\caption{Subtask 1B (Target Group) Dataset Distribution}
\label{tab:subtask1b-distribution}
\end{table}

\addtocounter{footnote}{-1}  
\footnotetext{\commonsplitfootnote}
Distribution analysis reveals consistent patterns between training and development sets, indicating proper data splitting. In particular, severe class imbalance poses challenges for model training, particularly for minority classes in Subtask 1A such as Sexism (0. 34\% in training) and Religious Hate (1.90\% in training). Similarly, in Subtask 1B, the None class constitutes 59.66\% of the training set, with minority classes like Society (6.21\%) and Community (7.42\%).

\section{Method Description}




\subsection{Adversarial Fine-tuning (FGSM)}
We use the Fast Gradient Sign Method (FGSM) \citep{goodfellow2014explaining} as a light-weight adversarial fine-tuning mechanism to improve robustness to small input perturbations (typos, obfuscation, or transliteration noise common in YouTube comments). FGSM perturbs token embeddings in the direction of the loss gradient:
\begin{equation}
\Delta = \epsilon \cdot \mathrm{sign}(\nabla_{\Theta} J(x,y;\Theta))
\end{equation}
and we minimize a combined loss:
\begin{equation}
\tilde{J}(x,y) = \alpha J(x,y) + (1-\alpha) J(x+\Delta,y).
\end{equation}
In our experiments we used $\epsilon=0.1$ and $\alpha=0.5$, applying FGSM on perturbed embeddings every other epoch during fine-tuning. We report these hyperparameters and ablate FGSM vs. standard fine-tuning in Section~5. FGSM is applied in embedding space only (not at token substitution level), which is computationally inexpensive compared to stronger iterative attacks.

While stronger iterative methods like Projected Gradient Descent (PGD)~\cite{geisler2024attacking} and Adversarial Weight Perturbations(AWP)~\cite{wu2020adversarial} could yield more robust perturbations, we selected FGSM for its single-step efficiency, reducing training time by ~50\% compared to PGD and ~57\% compared to AWP respectively(based on preliminary trials). 

\subsection{Data Augmentation}
To enhance our model's performance,we applied external augmentation by collecting multiple labeled hate speech samples from public datasets such as WoNBias \cite{aupi-etal-2025-wonbias} and Bangla Hate Speech Detection Dataset \cite{romim2021hate} for subtask 1A. These datasets were specifically selected to address categories like sexism, religious hate, and political hate. Initially, training on this combined dataset yielded improved accuracy on the development set. However, the same model performed poorly on the test set, likely due to domain mismatch or overfitting to the augmented data. As a result, we reverted to training our final models solely on the original training dataset.

\section{Results and Analysis}

We present comprehensive experimental results for both subtasks, demonstrating the effectiveness of our proposed ensemble-based approach with adversarial training and text normalization.

\subsection{Subtask 1A: Hate Speech Type Classification}

Table~\ref{tab:subtask1a-results} summarizes our experimental results for multi-class classification of Bengali text into six categories: Abusive, Sexism, Religious Hate, Political Hate, Profane, or None.

\begin{table}[h]
\centering
\footnotesize
\begin{tabular}{p{3.8cm}cc}
\hline
\textbf{Model} & \textbf{Dev} & \textbf{Test} \\
\hline
\multicolumn{3}{l}{\textit{DL Models}} \\
BiLSTM & 56.25 & 47.64 \\
LSTM+Attention & 55.18 & 54.41 \\
\hline
\multicolumn{3}{l}{\textit{Base LLMs}} \\
XLM-R-large & 72.81 & 70.54 \\
MuRIL-large & 71.02 & 70.29 \\
BanglaBERT & 70.74 & 70.31 \\
BanglaBERT-large & 70.51 & 68.13 \\
XLM-R-base & 70.50 & 70.15 \\
DistilBERT-multi & 68.03 & 68.49 \\
\hline
\multicolumn{3}{l}{\textit{K-Fold CV}} \\
MuRIL-large+KF & 73.61 & 71.88 \\
XLM-R-large+KF & 73.45 & 71.72 \\
BanglaBERT+KF & 73.29 & 72.05 \\
\hline
\multicolumn{3}{l}{\textit{K-Fold CV + Normalizer}} \\
BanglaBERT+N & 74.32 & 71.14 \\
XLM-R-large+N & 73.29 & 71.57 \\
MuRIL+N & 73.73 & 72.30 \\
\hline
\multicolumn{3}{l}{\textit{Adversarial attacks w/o Normalizer}} \\
BanglaBERT+KF+FGSM & 73.87 & 72.17 \\
MuRIL+KF+FGSM & 73.68 & 71.90 \\
\hline
\multicolumn{3}{l}{\textit{Adversarial Attacks w Normalizer}} \\
BanglaBERT+KF+FGSM+N & \textbf{74.88} & \textbf{72.33} \\
MuRIL+KF+FGSM+N & 73.81 & 71.31 \\
\hline
\end{tabular}
\caption{Subtask 1A Results (Micro F1 \%)}
\label{tab:subtask1a-results}
\end{table}

Our analysis shows several key findings: (1) Traditional deep learning models (BiLSTM, LSTM+Attention) significantly underperformed compared to transformer-based approaches, with F1 scores around 55-56\%. (2) Among base LLMs, XLM-R-large achieved the highest development performance (72.81\%), followed by MuRIL-large (71.02\%) and BanglaBERT (70.74\%). (3) K-fold cross-validation (KF) improved performance by 2-3\% across models, with MuRIL-large+KF reaching 73.61\% on Dev and 71.88\% on Test. (4) Text normalization (N) further boosted performance, particularly for BanglaBERT+N (74.32\% Dev, 71.14\% Test). (5) The combination of BanglaBERT with K-fold, FGSM adversarial training, and normalization (BanglaBERT+KF+FGSM+N) achieved the highest scores of 74.88\% on Dev and 72.33\% on Test, showing the effectiveness of combining these techniques.

\subsection{Subtask 1B: Target Group Classification}

Table~\ref{tab:subtask1b-results} presents results for hate speech target classification into four categories: Individuals, Organizations, Communities, or Society.

\begin{table}[h]
\centering
\footnotesize
\begin{tabular}{p{3.8cm}cc}
\hline
\textbf{Model} & \textbf{Dev} & \textbf{Test} \\
\hline
\multicolumn{3}{l}{\textit{Base LLMs}} \\
BanglaBERT & 72.09 & 70.25 \\
MuRIL-large & 71.93 & 70.93 \\
XLM-R-large & 71.38 & 71.23 \\
BanglaBERT-large & 69.90 & 68.88 \\
XLM-R-base & 71.09 & 70.19 \\
DistilBERT-multi & 69.27 & 68.46 \\
\hline
\multicolumn{3}{l}{\textit{K-Fold CV}} \\
MuRIL-large+KF & \textbf{74.96} & \textbf{73.44} \\
BanglaBERT+KF & 73.69 & 71.85 \\
XLM-R-large+KF & 71.53 & 68.07 \\
\hline
\multicolumn{3}{l}{\textit{K-Fold CV + Normalizer}} \\
BanglaBERT+N & 74.72 & 72.89 \\
MuRIL+N & 74.48 & 73.44 \\
XLM-R-large+N & 72.39 & 71.66 \\
\hline
\multicolumn{3}{l}{\textit{Adversarial attacks w/o Normalizer}} \\
BanglaBERT+KF+FGSM & 74.12 & 72.25 \\
MuRIL+KF+FGSM & 73.89 & 72.92 \\
XLM-R-large+KF+FGSM & 74.20 & 73.28 \\
\hline
\multicolumn{3}{l}{\textit{Adversarial attacks w Normalizer}} \\
BanglaBERT+KF+FGSM+N & 74.64 & 73.12 \\
MuRIL+KF+FGSM+N & 74.56 & 72.95 \\
XLM-R-large+KF+FGSM+N & 74.32 & 72.17 \\
\hline
\end{tabular}
\caption{Subtask 1B Results (Micro F1 \%)}
\label{tab:subtask1b-results}
\end{table}

For Subtask 1B, we observe: (1) Base transformer models showed consistent performance, with F1 scores ranging from 69.27\% to 72.09\% on Dev and 68.46\% to 71.23\% on Test. (2) MuRIL-large with K-fold cross-validation obtained the best scores of 74.96\% on Dev and 73.44\% on Test, indicating strong generalization. (3) Text normalization improved performance, with BanglaBERT+N and MuRIL+N reaching 74.72\% and 74.48\% on Dev, respectively, and 72.89\% and 73.44\% on Test. (4) Adversarial training with FGSM further enhanced results, with XLM-R-large with KFold CV  and FGSM attack achieving 74.20\% on Dev and 73.28\% on Test. (5) The combination of K-fold, FGSM, and normalization (e.g., BanglaBERT+KF+FGSM+N) yielded competitive performance (74.64\% Dev, 73.12\% Test), though slightly below MuRIL-large+KF on Test.

\subsection{Error Analysis}

\begin{figure}[t]
\centering
\includegraphics[width=\columnwidth]{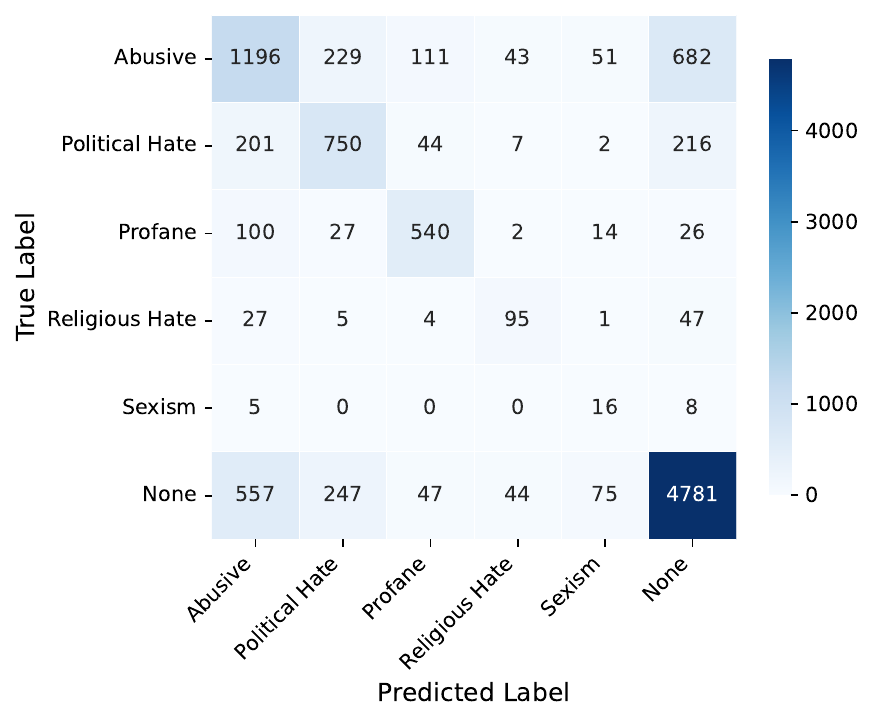}
\caption{Confusion matrix analysis for Subtask 1A using BanglaBERT with FGSM attack}
\label{fig:confusion-matrix}
\end{figure}

\begin{figure}[t]
\centering
\includegraphics[width=\columnwidth]{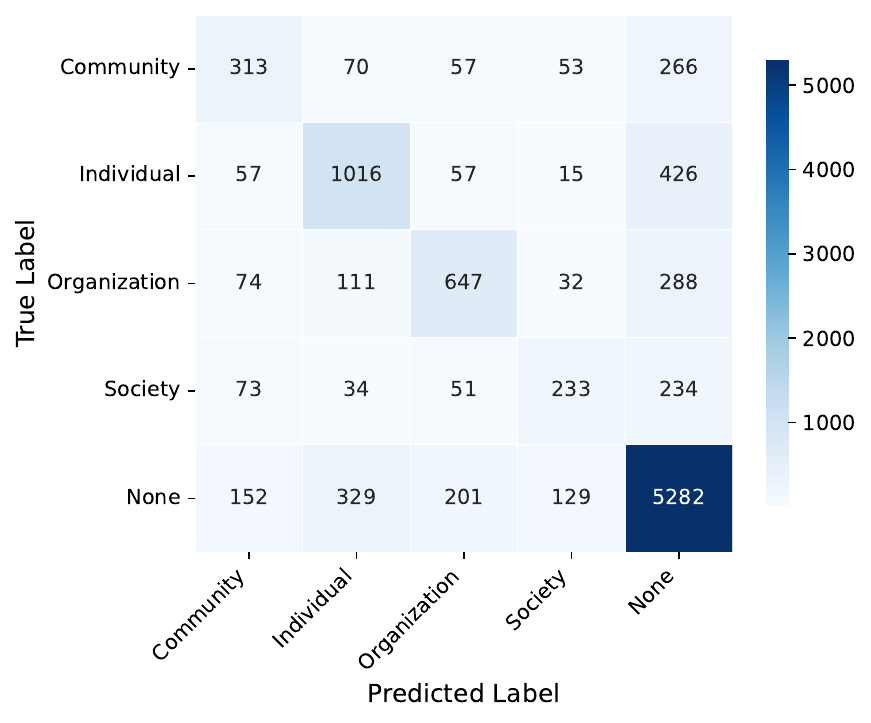}
\caption{Confusion matrix analysis for Subtask 1B using MuRiL with K Fold CV}
\label{fig:confusion-matrix-1b}
\end{figure}

Tables~\ref{tab:subtask1a-results} and~\ref{tab:subtask1b-results} show that BanglaBERT with FGSM adversarial training and text normalization obtains the best performance for Subtask 1A (72.33\% test F1), while MuRIL-large with K-fold achieved the best for Subtask 1B (73.44\% test F1). To gain further insight about system performance, the confusion matrices (Figures~\ref{fig:confusion-matrix} and~\ref{fig:confusion-matrix-1b}) are analyzed.

For Subtask 1A (Figure~\ref{fig:confusion-matrix}), the model obtains the highest True Positive Rate (TPR) of 83.13\% (4781/5751) for None category and 76.24\% (540/709) for Profane category. The model shows moderate performance for Political Hate with 61.48\% (750/1220) TPR and Religious Hate with 53.07\% (95/179) TPR. However, it provides the lowest TPR of 55.17\% (16/29) for Sexism category.

Our model significantly misclassified Abusive texts, where only 51.73\% (1196/2312) were correctly identified. A major portion of Abusive texts were misclassified as None (682 instances) and Political Hate (229 instances). This suggests the model struggles to distinguish between general abusive language and other hate categories, likely due to overlapping linguistic features such as aggressive vocabulary and derogatory terms. Political Hate texts also showed considerable confusion, with 216 instances misclassified as Abusive and 201 as None. The class imbalance, with Sexism representing only 0.34\% of training data, contributes to poor minority class performance.

For Subtask 1B (Figure~\ref{fig:confusion-matrix-1b}), the model achieves the highest TPR of 86.72\% (5282/6093) for None category and 64.59\% (1016/1571) for Individual targeting. The model shows moderate performance for Organization with 56.16\% (647/1152) TPR. However, it provides lower TPR for Community (41.24\%, 313/759) and Society (37.28\%, 233/625) categories.

The confusion between Individual and None categories shows 426 Individual texts misclassified as None, indicating difficulty identifying subtle personal attacks lacking explicit hate markers. Community-targeted hate shows significant confusion with None (266 misclassifications), while Society-targeted hate is frequently misclassified as None (234 instances). Organization texts are often confused with None (288 instances) and Individual (111 instances). This pattern suggests that abstract group-targeted criticism is difficult to distinguish from neutral text or personal attacks. The imbalanced dataset, where None represents 59.66\% of training data, biases toward the majority class, causing systematic misclassification of minority categories.

\subsection{Computational Analysis}
All experiments were conducted on dual NVIDIA T4 GPUs (T4x2 configuration) which were available through Kaggle and Google Colab platforms. Base LLM fine-tuning (e.g., BanglaBERT) required ~2-3 hours per epoch. The full pipeline with K-fold cross-validation (5 folds) and FGSM for BanglaBERT took approximately 6-7 hours in total. For larger models like MuRIL-large and XLM-R-large with K-fold and FGSM, training exceeded 12 hours, hitting the session limits on these platforms and requiring session restarts. FGSM added ~10-15\% overhead per epoch due to gradient computations but remained lightweight compared to iterative attacks like PGD.

\section{Conclusion}
\label{sec:bibtex}

Our experiments demonstrated the efficacy of integrating multilingual pre-trained models with robustness-enhancing techniques, especially in addressing class imbalances and noisy social media text from YouTube comments. The use of adversarial perturbations made the model more resilient to real-world changes like typos and informal language. Normalization, on the other hand, fixed script inconsistencies that are common in Bangla online content.

\section*{Limitations}
Our approach has some limitations even though the results are promising. The dataset has severe class imbalance, particularly for minority classes like "Sexism" and "Religious Hate" in subtask 1A, whereas we observed the same for "Community" and "Society" in subtask 1B. This might have led to under-performance on these categories, even with K-fold cross-validation and adversarial training. While we experimented with external data augmentation, it did not yield improvements on the test set, suggesting potential domain mis-matches between public datasets and the YouTube-specific corpus for the shared task.

Additionally, we intended to explore more advanced adversarial techniques, such as Geometry-Aware Adversarial Training (GAT)~\cite{zhu2022improving} and Adversarial Weight Perturbation (AWP)~\cite{wu2020adversarial}, to further enhance robustness of the models. However, these were infeasible due to computational constraints as fine-tuning larger models like XLM-RoBERTa-large and MuRIL-large uncased required approximately 5 hours per epoch. This exceeded the 12-hour session limits on platforms like Kaggle and Google Colab causing frequent restarts and complicating experimentation.

\bibliography{references}

\appendix

\end{document}